\documentclass{article}

\usepackage{arxiv}
\usepackage[utf8]{inputenc} % allow utf-8 input
\usepackage[T1]{fontenc}    % use 8-bit T1 fonts
\usepackage{hyperref}       % hyperlinks
\usepackage{url}            % simple URL typesetting
\usepackage{booktabs}       % professional-quality tables
\usepackage{amsfonts}       % blackboard math symbols
\usepackage{nicefrac}       % compact symbols for 1/2, etc.
\usepackage{microtype}      % microtypography
\usepackage{lipsum}         % Can be removed after putting your text content
\usepackage{graphicx}
\usepackage{subcaption}
\usepackage{amsmath}
\usepackage{cleveref}       % smart cross-referencing
\usepackage{pifont}
\usepackage[numbers,sort&compress]{natbib} 
\usepackage{doi}

\title{Noninvasive Intracranial Pressure Estimation Using Subspace System Identification and Bespoke Machine Learning Algorithms: A Learning-to-Rank Approach}

\newif\ifuniqueAffiliation
\uniqueAffiliationtrue

\ifuniqueAffiliation % Standard variant of author block
\author{ Anni Zhao\thanks{Use footnote for providing further
		information about author (webpage, alternative
		address)---\emph{not} for acknowledging funding agencies.} \\
	Department of Computer Science\\
	Cranberry-Lemon University\\
	Pittsburgh, PA 15213 \\
	\texttt{hippo@cs.cranberry-lemon.edu} \\
	\And
	\href{https://orcid.org/0000-0000-0000-0000}{\includegraphics[scale=0.06]{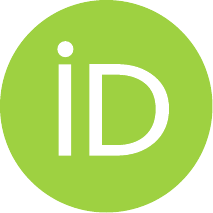}\hspace{1mm}Elias D.~Striatum} \\
	Department of Electrical Engineering\\
	Mount-Sheikh University\\
	Santa Narimana, Levand \\
	\texttt{stariate@ee.mount-sheikh.edu} \\
}

\author{\hspace{1mm}Anni Zhao\\Center for Data Science\\
    Emory University\\
    Atlanta, GA 30322, USA\\
    \texttt{anni.zhao@emory.edu}
    \And
    \hspace{1mm}Ayca Ermis\\
    Center for Data Science\\
    Emory University\\
    Atlanta, GA 30322, USA\\
    \texttt{ayca.ermis@emory.edu}
    \And
    Jeffrey Robert Vitt\\
    University of California, Davis\\
    Sacramento, CA 95816, USA\\
    \texttt{jrvitt@ucdavis.edu}
    \And
    Sergio Brasil\\
    S\~{a}o Paulo University\\
    S\~{a}o Paulo, Brazil\\
    \texttt{sbrasil@alumni.usp.br}
    \And
    Wellingson Paiva\\
    S\~{a}o Paulo University\\
    S\~{a}o Paulo, Brazil\\
    \texttt{wellingsonpaiva@yahoo.com.br}
    \And
    Magdalena Kasprowicz\\
    Wroclaw University of Science and Technology\\
    Wroclaw 50-370, Poland\\
    \texttt{magdalena.kasprowicz@pwr.edu.pl}
    \And
    Malgorzata Burzynska\\
    Wroclaw Medical University\\
    Wroclaw 50-556, Poland\\
    \texttt{malgorzata.burzynska@umw.edu.pl}
    \And
    Robert Hamilton\\
    NeuraSignal\\
    Los Angeles, CA 90024, USA\\
    \texttt{robert@neurasignal.com}
    \And
    Runze Yan\\
    Emory University\\
    Atlanta, GA 30322, USA\\
    \texttt{runze.yan@emory.edu}
    \And
    Ofer Sadan\\
    Emory University\\
    Atlanta, GA 30322, USA\\
    \texttt{ofer.sadan@emory.edu}
    \And
    J.\ Claude Hemphill\\
    University of California, San Francisco\\
    San Francisco, CA 94110, USA\\
    \texttt{claude.hemphill@ucsf.edu}
    \And
    Lieven Vandenberghe\\
    University of California, Los Angeles\\
    Los Angeles, CA 90095, USA\\
    \texttt{vandenbe@ucla.edu}
    \And
    Xiao Hu\thanks{Corresponding author: Xiao Hu.} \\
    Emory University\\
    Atlanta, GA 30322, USA\\
    \texttt{xiao.hu@emory.edu}
}
\else
\usepackage{authblk}

\newbox{\orcid}\sbox{\orcid}{\includegraphics[scale=0.06]{orcid.pdf}} 
\author[1]{%
	\href{https://orcid.org/0000-0000-0000-0000}{\usebox{\orcid}\hspace{1mm}Anni Zhao\thanks{\texttt{anni.zhao@emory.edu}}}%
}
\author[1]{%
	\href{https://orcid.org/0000-0000-0000-0000}{\usebox{\orcid}\hspace{1mm}Ayca Ermis\thanks{\texttt{stariate@ee.mount-sheikh.edu}}}%
}
\author[3]{%
	\href{https://orcid.org/0000-0000-0000-0000}{\usebox{\orcid}\hspace{1mm}Jeffrey Robert Vitt\thanks{\texttt{jrvitt@ucdavis.edu}}}%
}
\author[4]{%
	\href{https://orcid.org/0000-0000-0000-0000}{\usebox{\orcid}\hspace{1mm}Sergio Brasil\thanks{\texttt{sbrasil@alumni.usp.br}}}%
}
\author[4]{%
	\href{https://orcid.org/0000-0000-0000-0000}{\usebox{\orcid}\hspace{1mm}Wellingson Paiva\thanks{\texttt{wellingsonpaiva@yahoo.com.br}}}%
}
\author[5]{%
	\href{https://orcid.org/0000-0000-0000-0000}{\usebox{\orcid}\hspace{1mm}Magdalena Kasprowicz\thanks{\texttt{magdalena.kasprowicz@pwr.edu.pl}}}%
}
\author[6]{%
	\href{https://orcid.org/0000-0000-0000-0000}{\usebox{\orcid}\hspace{1mm}Malgorzata Burzynska\thanks{\texttt{malgo-
rzata.burzynska@umw.edu.pl}}}%
}
\author[7]{%
	\href{https://orcid.org/0000-0000-0000-0000}{\usebox{\orcid}\hspace{1mm}Robert Hamilton\thanks{\texttt{robert@neurasignal.com}}}%
}
\author[8]{%
	\href{https://orcid.org/0000-0000-0000-0000}{\usebox{\orcid}\hspace{1mm}Runze Yan\thanks{\texttt{runze.yan@emory.edu}}}%
}
\author[9]{%
	\href{https://orcid.org/0000-0000-0000-0000}{\usebox{\orcid}\hspace{1mm}Ofer Sadan\thanks{\texttt{ofer.sadan@emory.edu}}}%
}
\author[10]{%
	\href{https://orcid.org/0000-0000-0000-0000}{\usebox{\orcid}\hspace{1mm}J. Claude Hemphill\thanks{\texttt{claude.hemphill@ucsf.edu}}}%
}
\author[11]{%
	\href{https://orcid.org/0000-0000-0000-0000}{\usebox{\orcid}\hspace{1mm}Lieven
Vandenberghe\thanks{\texttt{vandenbe@ucla.edu}}}%
}
\affil[1]{Department of Computer Science, Cranberry-Lemon University, Pittsburgh, PA 15213}
\affil[2]{Department of Electrical Engineering, Mount-Sheikh University, Santa Narimana, Levand}
\fi

\renewcommand{\shorttitle}{\textit{arXiv} Template}

\hypersetup{
pdftitle={Noninvasive Intracranial Pressure Estimation Using Subspace System Identification and Bespoke Machine Learning Algorithms: A Learning-to-Rank Approach},
pdfsubject={q-bio.NC, q-bio.QM},
pdfauthor={David S.~Hippocampus, Elias D.~Striatum},
pdfkeywords={First keyword, Second keyword, More},
}

\begin{document}
\maketitle

\begin{abstract}
	\textit{Objective:}  Accurate noninvasive estimation of intracranial pressure (ICP) remains a major challenge in critical care. We developed a bespoke machine learning algorithm that integrates system identification and ranking-constrained optimization to estimate mean ICP from noninvasive signals. \textit{Methods:} A machine learning framework was proposed to obtain accurate mean ICP values using arbitrary noninvasive signals. The subspace system identification algorithm is employed to identify cerebral hemodynamics models for ICP simulation using arterial blood pressure (ABP), cerebral blood velocity (CBv), and R-wave to R-wave interval (R-R interval) signals in a comprehensive database. A mapping function to describe the relationship between the features of noninvasive signals and the estimation errors is learned using innovative ranking constraints through convex optimization. Patients across multiple clinical settings were randomly split into testing and training datasets for performance evaluation of the mapping function. \textit{Results:} The results indicate that about 31.88\% of testing entries achieved estimation errors within 2 mmHg and 34.07\% of testing entries between 2 mmHg to 6 mmHg from the nonlinear mapping with constraints. \textit{Conclusion:} Our results demonstrate the feasibility of the proposed noninvasive ICP estimation approach. \textit{Significance:} Further validation and technical refinement are required before clinical deployment, but this work lays the foundation for safe and broadly accessible ICP monitoring in patients with acute brain injury and related conditions.
\end{abstract}

% keywords can be removed

\section{Introduction}
Intracranial pressure (ICP) monitoring is invaluable for preventing secondary brain injury in acute brain injury (ABI), including traumatic brain injury (TBI), subarachnoid hemorrhage (SAH), or intracerebral hemorrhage (ICH), and is also indicated in other conditions with suspected intracranial hypertension (IH) (e.g., pseudotumor cerebri). However, the invasive methods for ICP measurement carry the risks of procedural complications and infection, thereby limiting more widespread use \cite{GunnHu2020}, \cite{CarneyTotten2017}, \cite{ HawrylukRubiano2020}. Consequently, ICP monitoring is confined to a narrow range of acute neurologic conditions in the critical care setting \cite{BrattonChestnut2007}. 

The ongoing development of several accurate and clinically viable noninvasive ICP (nICP) techniques reflect the need for such methods \cite{CanacJalaleddini2020}. Despite non-invasive devices aim to monitor different brain physiological phenomena and help detect IH, none of the currently available techniques can provide continuous ICP estimation at the bedside. Such methods include visual evoked potentials \cite{delRedondo2016}, displacement of the skull \cite{KomutKozac2016}, brain tissue resonance \cite{MotschmannMuller2001}, cerebrospinal fluid (CSF) flow and cerebral blood flow (CBF) assessed by phase-contrast magnetic resonance imaging (MRI) \cite{CzarnikGawda2007}, cerebral blood velocity (CBv) as measured by Transcranial Doppler Ultrasound (TCD)\cite{WeiKrakauskaite2025}, \cite{RasuloCalza2022}, \cite{PaneraiBrassard2023}, the tympanic membrane displacement \cite{UenoBallard2003}, and electroencephalography (EEG) \cite{PoseVidela2025}, acoustic signals through sound propagation from the ear \cite{GanslandtMourtzoukos2017, HerklotsMoudrous2017}, automated pupillometry \cite{OddoTaccone2023} and optic nerve sheath diameter ultrasound \cite{AletrebyAlharthy2022, MartinezVasquez2024}. More recently, the Brain4Care (B4C) system was developed by capturing micrometric skull displacements caused by pulsatile changes in ICP \cite{FrigieriBrasil2025, FrigieriGoncalves2025}. Therefore, this system records skull pulsatile waveforms that have been compared and validated using invasive ICP waveforms \cite{UysalWilliams2025}, \cite{BrasilSolla2025}, \cite{deRocha2025}.

In recent decades, model-based algorithms have been developed using hemodynamic models and various physiological signals \cite{KashifVerghese2012, ImaduddinFanelli2019}. A fully automated, real-time, and calibration-free algorithm has been proposed for pediatric intracranial pressure (ICP) estimation \cite{FanelliVonberg2019}. In related work, a model-based spectral methodology was developed for noninvasive ICP estimation \cite{JaishankarFanelli2019}. A transfer function based algorithm is adopted to estimate the intracranial pressure noninvasively for idiopathic normal pressure hydrocephalus patients \cite{EvensenORourke2018}. These approaches typically rely on predefined mathematical representations of the cardiovascular and cerebrospinal systems to estimate ICP. While they offer interpretability and are grounded in physiological understanding, their accuracy can be limited by model assumptions and the quality of input signals. To address these limitations, data-driven methods have gained increasing attention. Among them, machine learning has emerged as a promising alternative for nICP estimation, capable of capturing complex, nonlinear relationships that traditional models may overlook \cite{MegjhaniTerilli2023}. In particular, photoplethysmography (PPG) was introduced to assess its potential for nICP estimation using a range of machine learning algorithms \cite{BradleyKyriacou2024}. A systematic review for using machine learning approaches to intracranial pressure prediction can be found in \cite{BradleyRoldan2023}. 

In this study, we present a novel framework that integrates system identification with ranking-constrained machine learning for individualized nICP calibration using multimodal noninvasive signals. By leveraging a multicenter database and comprehensive feature extraction, our approach seeks to advance algorithmic solutions for noninvasive ICP estimation. Unlike conventional methods, it leverages mapping-function learning with problem-specific information and a bespoke machine learning algorithm to overcome past limitations and improve accuracy when selecting optimal dynamic models.

The remainder of this paper is organized as follows. Section \ref{sec:algorithm} describes the proposed algorithm for noninvasive intracranial pressure estimation. Section \ref{sec:dataprocessing} presents the data processing for the physiological signals. Section \ref{sec:results} presents the experimental results and performance evaluation of the nICP system. Section \ref{sec:discussion} and \ref{sec:conclusion} provide future research directions and conclusions.

\section{Algorithm}
\label{sec:algorithm}

This section introduces the proposed algorithm for nICP estimation. Section \ref{SsecNICPFramework} proposes the framework for noninvasive intracranial pressure estimation. Section \ref{SsLDM} provides a brief introduction of the subspace system identification algorithm. Section \ref{SsecLearnMapplingFunction} presents the theoretical formulation and an approximate algorithm to learn mapping-functions. Section \ref{SsecNoninvasiveFeatures} describes the procedures for noninvasive feature extraction.

\subsection{Noninvasive Intracranial Pressure Calibration Framework}
\label{SsecNICPFramework}
\begin{figure*}[ht]
	\centering
	\includegraphics[width=0.8\textwidth]{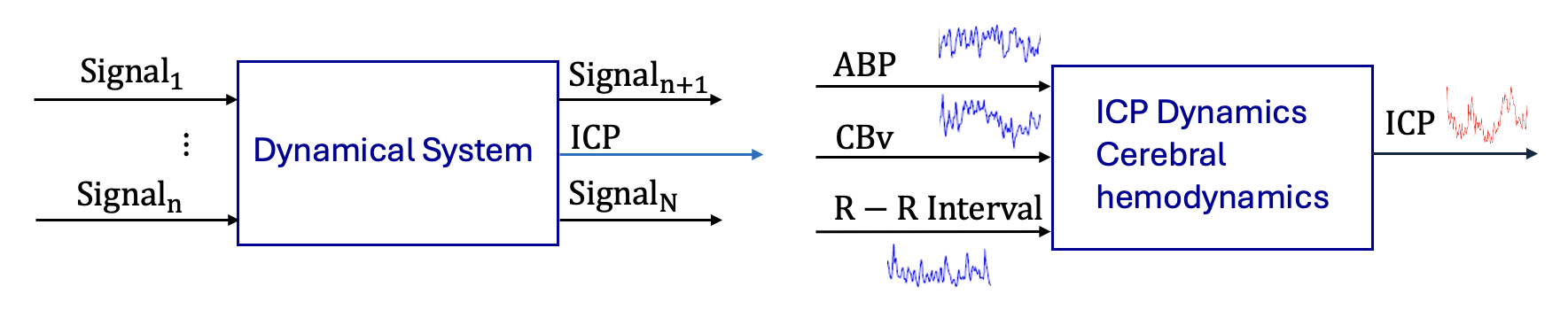} % Adjust width as needed
	\caption{Left: A dynamical system view of ICP and noninvasive signals. Right: A dynamical system view of ICP, ABP, CBv, and R-R interval.}
	\label{FigZhao1}
\end{figure*}

From the perspective of dynamic system modeling, the underpinnings of the proposed nICP framework can be understood by analyzing the possible forms of relationships among continuous ICP and noninvasive signals. We adopted the view presented in Fig. \ref{FigZhao1} for our nICP framework. In this framework, multiple noninvasive signals, $Signal_{i}, i\in[1,2,\cdots,N]$, are related to ICP via a dynamical system with ICP being one of the output signals. This concept applies to our choice of ABP and CBv for estimating ICP, because it is physiologically plausible that ABP and the R–R interval serve as two inputs to the coupled CBF and CSF circulatory systems, from which both ICP and CBv are outputs. ABP and CBv affect the brain hemodynamics through cerebral perfusion pressure and autoregulation. R-R interval influences the pulsatility of cerebral blood flow and autoregulation interactions. The dynamic system view of ABP, CBv, R-wave to R-wave interval (R-R interval), and ICP is shown on the right-hand side in Fig. \ref{FigZhao1}. 

\begin{figure*}[ht]
	\centering
	\includegraphics[width=0.85\textwidth]{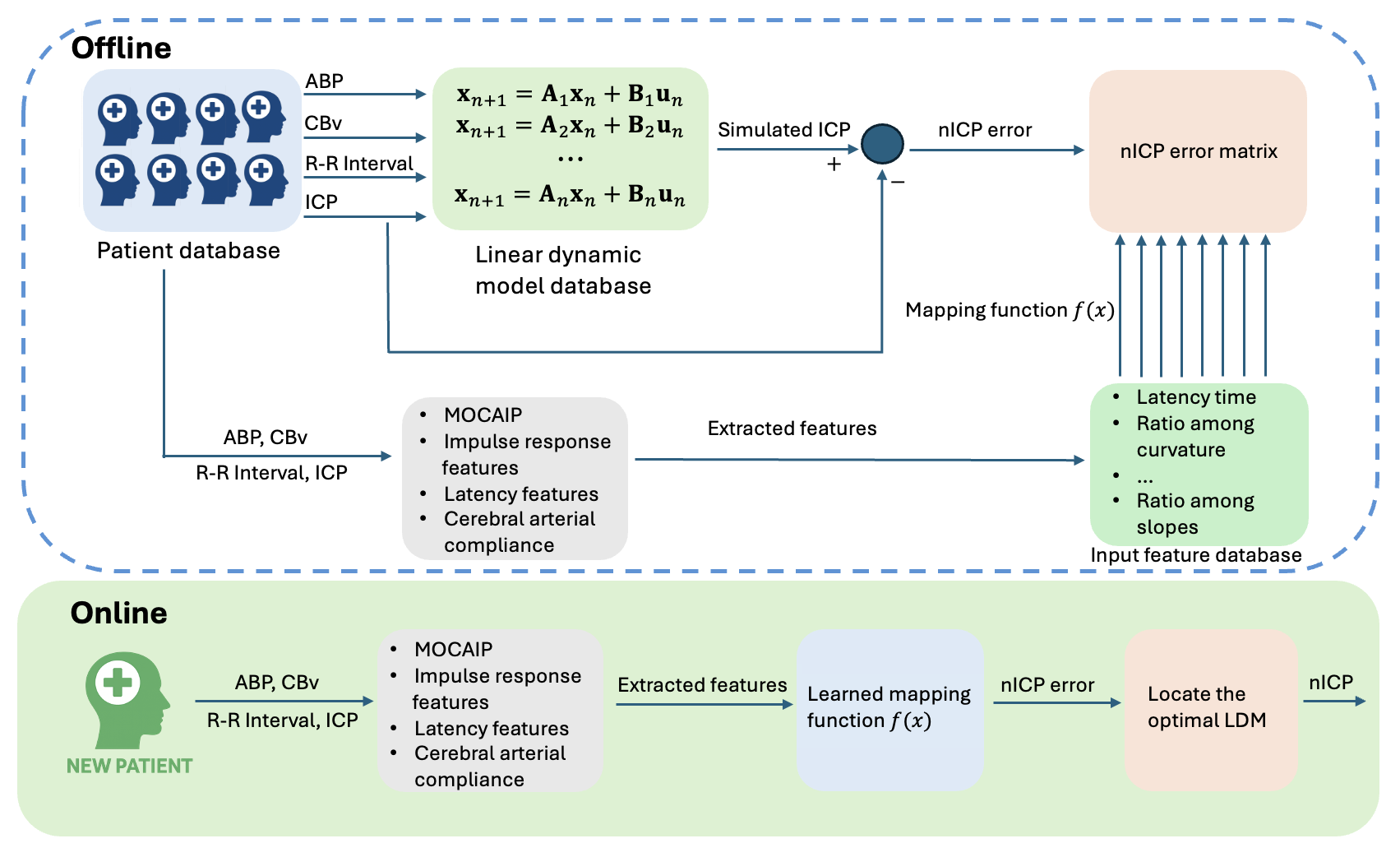} % Adjust width as needed
	\caption{Block diagram of the nICP calibration framework. The contents within the blue dashed line represent the offline training procedure. The green area represents the online training procedure.}
	\label{FigZhao2}
\end{figure*}

The central idea behind our approach is to first build a collection of dynamic models of the dynamical system and then use an instance of ICP-related signals to locate the optimal model within the collection to simulate ICP for this particular instance. The algorithm framework to realize this idea is shown in Fig. \ref{FigZhao2}. It consists of an offline training and an online implementation section. In offline training, instances of invasive ICP, ABP, CBv, and R-R interval signals are analyzed to yield three databases: 1) linear dynamic models (LDMs), 2) noninvasive features, and 3) mapping-functions. The LDM database contains the input/output models relating ABP, CBV, and R-R interval to ICP built from each signal instance. The feature database contains feature vectors, each of which is extracted from an instance of ABP, CBv, and R-R interval. These feature vectors are used as inputs to the mapping-functions. The mapping-function database contains a mapping-function for each entry in the LDM database to estimate nICP error that would result from using the corresponding LDM on a given instance of ABP, CBv, and R-R interval. The outputs from the mapping-functions would be the nICP estimation errors from the LDMs.
To estimate ICP, a new instance of noninvasive signals is first analyzed to extract a feature vector that is further evaluated by each mapping-function in the database. The output from the $i$-th mapping-function is the estimated nICP error using the $i$-th model in the database. The estimated nICP errors from all mapping-functions are then ranked so that the LDM with the smallest error will be chosen to simulate nICP. In the following subsections \ref{SsLDM}, \ref{SsecLearnMapplingFunction}, and \ref{SsecNoninvasiveFeatures}, the algorithms used to build linear dynamic models, learn mapping-functions, and extract noninvasive features are explained in detail. 

\subsection{Linear Dynamic Models}
\label{SsLDM}
The subspace system identification algorithm is adopted to identify the dynamical system with ABP, CBv, R-R interval as inputs and ICP as output. The structure of the dynamical system is shown in Fig. \ref{FigZhao1}. The subspace system identification algorithm is a data-driven algorithm from the control theory for state-space model identification \cite{Katayama2005}. The identified system can be written in state-space form as,
\begin{align}
	\mathbf{x}_{n+1} &= \mathbf{A}\mathbf{x}_n + \mathbf{B}\mathbf{u}_n + \mathbf{w}_n\\
	y_n &= \mathbf{C}\mathbf{x}_n + \mathbf{D}\mathbf{u}_n + v_n
\end{align} 
where $\mathbf{u}_n = [x_{ABP}, x_{CBv}, x_{R-R\hspace{0.5em}interval}]^T\in\mathbb{R}^{3\times1}$ represents the ABP, CBv, and R-R interval signals, $y_n = x_{ICP}$ represents the ICP signal. $\mathbf{A}$, $\mathbf{B}$, $\mathbf{C}$, and $\mathbf{D}$ are the system matrices to be estimated. $\mathbf{w}_n$ is the process noise, $v_n$ is the measurement noise. Given the inputs and outputs of the system, the subspace system identification algorithm estimates the system matrices by making use of the system inputs $\mathbf{u}_n$ and output $y_n$. Here, we adopted the optimal model order identified by the MATLAB System Identification Toolbox for each model.

\subsection{Mapping Functions Learning}
\label{SsecLearnMapplingFunction}
Assume $N$ signal entries are collected as the training data. A single entry is defined as an individual signal segment selected from each recording, with a length of 360 beats. Details can be found in Section \ref{sec:dataprocessing}. From these $N$ signal entries, we identify $N$ LDMs and extract $N$ feature vectors. To learn mapping functions, $N\times N$ model simulations are first performed, where each  LDM is used to simulate ICP for each signal entry. By comparing simulated and measured ICP for each entry, we obtain an $N\times N$ error matrix $\mathbf{E}$. $\mathbf{E}$ is organized such that its $i$-th row contains errors for $i$-th entry using $N$ LDMs and the $i$-th column contains errors from using the $i$-th LDM to simulate ICP for the $N$ entries. 
Beginning with using a linear mapping-function for the $i$-th LDM, we can estimate nICP error for a given noninvasive feature vector $\mathbf{f}\in\mathbb{R}^{d\times1}$ as $e=\mathbf{f}^T \mathbf{b}_i$ where $\mathbf{b}_i\in\mathbb{R}^{d\times1}$ is the coefficient vector for this mapping-function. To estimate $\mathbf{b}_i,i=1,\cdots,N$ using error matrix $\mathbf{E}$ and feature matrix $\mathbf{F}$ from the training data, we can formulate the problem:
\begin{equation}
    \min_{\substack{\mathbf{b}_i}}||\mathbf{F}\mathbf{b}_i - \mathbf{e}_{:,i}||_2, i = 1, \cdots, N\label{EqLM_Wo_Constraints}
\end{equation}
where $\mathbf{F}\in\mathbb{R}^{N\times d}$, the $i$-th row of feature matrix $\mathbf{F}$ is the feature vector from $i$-th entry. $\mathbf{e}_{:,i}\in\mathbb{R}^{N\times 1}$ is the column vector of error matrix $\mathbf{E}\in\mathbb{R}^{N\times N}$. $||\cdot||_{2}$ is the $L_2$ norm of a matrix. In essence, we solve $N$ independent least squares problems. The original formulation in Eq. (\ref{EqLM_Wo_Constraints}) is hereafter referred to as `Linear Mapping without Constraints' for illustration in the following sections. 

\begin{figure*}[ht]
	\centering
	\includegraphics[width=0.85\textwidth]{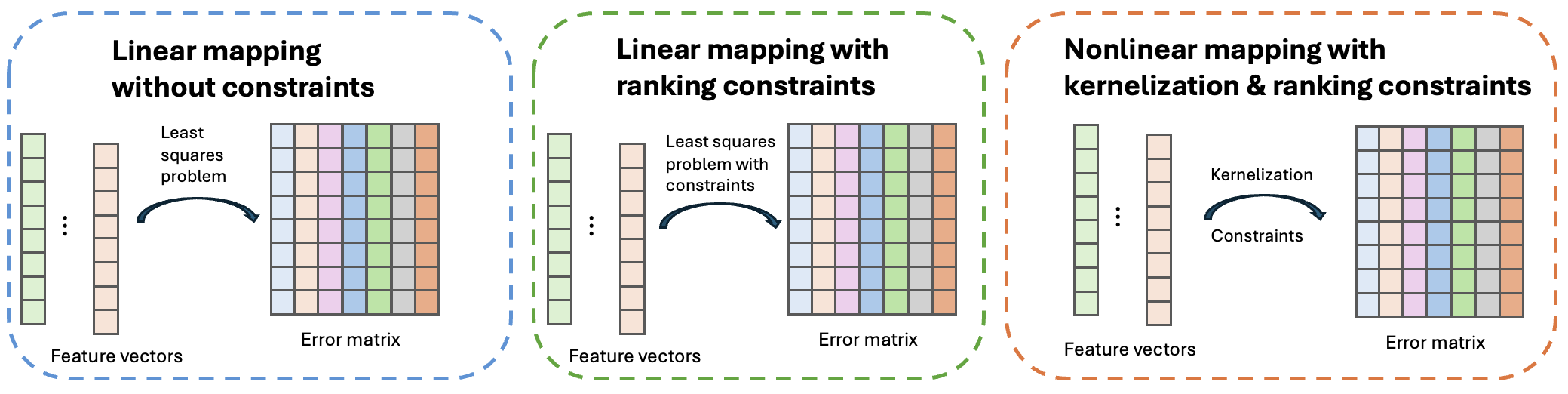} % Adjust width as needed
	\caption{Summary of various algorithms to learn the mapping function. Left: Linear mapping function learning without constraints. Middle: Linear mapping function learning with constraints. Right: Nonlinear mapping with kernelization \& ranking constraints.}
	%\label{FigLearnMappingFunction}
    \label{FigZhao3}
\end{figure*}

However, this simple formulation misses an important characteristic that $\mathbf{b}_i$ should have. The order of $e_{i,j}, i, j=1,\cdots,N$ is important because it tells the ranking of LDM models with regard to the accuracy of their nICP estimates for $j$-th database entry. Therefore, it would be beneficial to learn $\mathbf{b}_i$s in such a way that $\mathbf{f}_i^T \mathbf{b}_k$ and $\mathbf{f}_i^T \mathbf{b}_l$ obeys the ranking between $e_{i,k}$ and $e_{i,l}$. Obviously, independent learning of each $\mathbf{b}_i$ cannot ensure this important ranking relationship. To incorporate the ranking constraints, we formulate the problem as:
\begin{align}
	\min_{\substack{\mathbf{b}}_i}||\mathbf{F}\mathbf{b}_i - \mathbf{e}_{:,i}||_{2} &+ \gamma||\boldsymbol{\sigma}||_2, i = 1, \cdots, N\\
	subject\hspace{0.5em}to\hspace{0.5em}&\sigma_j \geq 0\\
	\boldsymbol{\alpha}_j^T\mathbf{F}\mathbf{B}\boldsymbol{\beta}_j \leq \sigma_j, j &= 1,2,\cdots,N\times\binom{N}{2}
\end{align}
where $\mathbf{B}\in\mathbb{R}^{d\times N}$ is a matrix with  $\mathbf{b}_i$  being the $i$-th column, $\boldsymbol{\alpha}_j$  is a $N\times1$ vector with only one nonzero element equal to 1, $\boldsymbol{\beta}_j$ is a $N\times1$ vector with only two nonzero elements equal to 1 and -1, respectively. For example, $\boldsymbol{\alpha}_j$ could be in the form as $[0 \cdots 1 \cdots 0]^T$, $\boldsymbol{\beta}_j$ could be in the form as $[0 \cdots 1 \cdots -1 \cdots 0]^T$. So $\boldsymbol{\alpha}_j^T\mathbf{F}\mathbf{B}\boldsymbol{\beta}_j=\boldsymbol{f}_m^T (\boldsymbol{b}_k-\boldsymbol{b}_l)$ expresses one ranking condition for an arbitrary $m$, $k$, $l$ depending on the locations of those nonzero elements in $\boldsymbol{\alpha}_j$ and $\boldsymbol{\beta}_j$. Here $m$ is the index for the entry. $\sigma_j$ is a slack variable for relaxing the constraints so that $\boldsymbol{f}_m^T (\boldsymbol{b}_k-\boldsymbol{b}_l)<0$ does not need to be strictly satisfied. In the above problem formulation, $\gamma$ is a hyperparameter that controls the tradeoff between the error of estimating nICP error and the degree of constraint violations. Importantly, the problem is convex, and a global optimal solution exists. However, this is a large-scale problem given the huge number, $N\times\binom{N}{2}$, of constraints because the number of database entries $N$ needs to be big for achieving diversity in the dynamics of stored signals and ensure wide applicability of the resultant nICP system. We also developed an approximate algorithm for linear mapping to speed up the computation and a nonlinear mapping algorithm with kernelization to improve the estimation accuracy. The illustration of various algorithms can be found in Fig. \ref{FigZhao3}. 

\subsection{An Approximate Algorithm to Learn Mapping Functions}
\label{SssecApproximatedMappingFuc}
Instead of enforcing all constraints at the same time, we will sequentially solve N sub-problems, the $n$-th of which will have the following form
\begin{align}
	\min_{\substack{\mathbf{b}_n}}||\mathbf{F}\mathbf{b}_n - \mathbf{e}_{:,n}||_2 + \gamma||&\boldsymbol{\sigma}||_2+\rho||\mathbf{b}_n||_2\\
	subject\hspace{0.5em}to\hspace{0.5em}\sigma_l \geq 0,l = 1,&\cdots, 2N\\
	\mathbf{f}_j^T\mathbf{b}_n < min(\hat{e}_{j, L^{-1}(j,n)})\hspace{0.5em}+\hspace{0.5em} &\sigma_j\\
	\mathbf{f}_j^T\mathbf{b}_n > max(\hat{e}_{j, L^{+}(j,n)})\hspace{0.5em}+\hspace{0.5em} &\sigma_{j+N}, j = 1,\cdots, N\label{Equation10}
\end{align}
In this formulation, we assume we have estimated $\mathbf{b}_i, i=1,\cdots, n-1$ before solving $\mathbf{b}_n$. The incorporation of ranking constraints is done by using two lists:
\begin{itemize}
	\item $L^+(j,n) = \{k|e_{j,n} > e_{j,k}\hspace{0.5em}\&\hspace{0.5em}k = 1, \cdots, n-1\}$, which collects indices of those $e_{j,k}$ that are smaller than $e_{j,n}$.
	\item $L^-(j,n) = \{k|e_{j,n} < e_{j,k}\hspace{0.5em}\&\hspace{0.5em}k = 1, \cdots, n-1\}$, which collects indices of those $e_{j,k}$ that are greater than $e_{j,n}$.
\end{itemize}
Further let $\hat{e}_{j,i} = \mathbf{f}^T_j\mathbf{\hat{b}}_i$ be the estimate of $e_{j,i}$ where $\mathbf{\hat{b}}_i$ is the estimated $\mathbf{b}_i$. Then, we can see that $\mathbf{f}^T_j\mathbf{b}_n$ needs to be greater than all $\hat{e}_{j,k}$ with $k$ in list $L^+(j,n)$. This condition can be simplified as $\mathbf{f}^T_j\mathbf{b}_n > max(\hat{e}_{j,L^+(j,n)})$. Similarity, $\mathbf{f}^T_j\mathbf{b}_n$ needs to be smaller than all $\hat{e}_{j,k}$ with $k$ in list $L^{-1}(j,n)$ that leads to $\mathbf{f}^T_j\mathbf{b}_n < min(\hat{e}_{j,L^-(j,n)})$.
Given the above derivation, the maximal number of constraints is $2N$ at each step. When conflicts occur where $min(\hat{e}_{j,L^-(j,n)}) < max(\hat{e}_{j,L^+(j,n)})$, we randomly keep only one of the two constraints. Also, slack variable $\boldsymbol{\sigma}$ is introduced as well to relax the constraints as a usual practice. Here, a regularization term $\rho||\mathbf{b}_n||_2$ is introduced to deal with the under-determined system when the feature dimension is higher than the number of samples. The approximation algorithm with ranking constraints formulated above is hereafter referred to as `Linear Mapping with Constraints' for illustration in the paper.

\subsection{Kernelization to Achieve Nonlinear Mapping Functions}
\label{ssseionKernelizedNM}
To further enhance the flexibility of the mapping-function, we demonstrate (after dropping all subscripts and introducing a L2 regularization term) that the above linear solution can be expressed solely in the form of inner products among various feature vectors and hence a ‘‘kernel trick’’ can be played to replace those inner products with a nonlinear kernel function. In this way, we obtain a nonlinear mapping that takes into consideration the ranking constraints. This process can be compactly expressed as:

Step 1): re-formulation in standard form
\begin{align}
	\min_{\substack{\mathbf{b}_i}}&||\mathbf{F}\mathbf{b}_i-\mathbf{e}_{:,i}||_2+\gamma||\boldsymbol{\sigma}||_2+\rho||\mathbf{b}_i||_2\\
	subject\hspace{0.5em}to\hspace{0.5em}&\sigma_j\geq0\hspace{0.5em}\&\hspace{0.5em}\mathbf{f}_j^T\mathbf{b}_i\leq \omega_j +\sigma_j, j = 1, \cdots, N
\end{align}

Step 2): Assume that $\mathbf{b}_i = \mathbf{F}^T\mathbf{u}$, then solution must be in row space of $\mathbf{F}$
\begin{align}
	\min_{\substack{\mathbf{u}}}&||\mathbf{F}\mathbf{F}^T \mathbf{u}-\mathbf{e}_{:,i}||_2+\gamma||\boldsymbol{\sigma}||_2+\rho \mathbf{u}^T\mathbf{F}\mathbf{F}^T\mathbf{u}\\
	subject\hspace{0.5em}to\hspace{0.5em}&\sigma_j\geq0\hspace{0.5em}\&\hspace{0.5em}\mathbf{f}_j^T\mathbf{F}^T\mathbf{u}\leq \omega_j+\sigma_j, j = 1, \cdots, N
\end{align}

Step 3): substitution of $\mathbf{z} = \mathbf{F}\mathbf{F}^T\mathbf{u}$
\begin{align}
	\min_{\substack{\mathbf{z}}}&||\mathbf{z}-\mathbf{e}_{:,i}||_2+\gamma||\boldsymbol{\sigma}||_2+\rho \mathbf{z}^T(\mathbf{F}\mathbf{F}^T)^{-1}\mathbf{z}\\
	subject\hspace{0.5em}to\hspace{0.5em}&\sigma_j\geq0\hspace{0.5em}\&\hspace{0.5em}z_j \leq \omega_j+\sigma_j, j = 1, \cdots, N
\end{align}

where $\mathbf{u}\in\mathbb{R}^{N\times 1}$, $\mathbf{z}\in\mathbb{R}^{N\times 1}$. $w_j$ is the ranking constraints from section \ref{SssecApproximatedMappingFuc}. $\rho$ and $\gamma$ are the hyperparameters that controls the trade-off between different terms in the objective function. After estimating $\mathbf{z}$ as the solution to the problem formulated at Step 3), an estimate of nICP error for a new feature vector $\mathbf{f}_{N+1}$ can be obtained as $e_{N+1}=\mathbf{f}_{N+1}^T \mathbf{F}^T (\mathbf{F}\mathbf{F}^T)^{-1}\mathbf{z}$ where only inner products among rows of $\mathbf{F}$ and between rows of $\mathbf{F}$ and $\mathbf{f}_{N+1}$ are involved. By replacing inner products with kernel functions $k(x,y)$, we have $e(\mathbf{f}_{N+1})=\sum_{i=1}^{N}k(\mathbf{f}_{N+1}^T,\mathbf{f}_{i,:})\mathbf{K}_{i,:}^{-1}\mathbf{z}$ where $\mathbf{K}^{-1}$ is the inverse of kernel matrix $\mathbf{K}$ of training features with $\mathbf{K}_{i,j}=\mathbf{k}(\mathbf{f}_{i,:},\mathbf{f}_{j,:})$. Here Gaussian function and polynominal function are adopted as the kernel function $\mathbf{k}(\mathbf{x},\mathbf{y})$ and the regularization term $\rho||\mathbf{b}_i||_2$ is introduced for the nonlinear mapping function to deal with the under-determined system when the feature dimension is higher than the number of samples. 

The Gaussian kernel function is defined as,
\begin{equation}
	\mathbf{k}(\mathbf{x},\mathbf{y}) = e^{-\frac{|\mathbf{x}-\mathbf{y}|^2}{2t^2}}
\end{equation}
where $|\mathbf{x}-\mathbf{y}|^2$ stands for the Euclidean distance squared between the two feature vectors, $t$ stands for the smoothing parameter of the kernel function. Here we adopted the median value of the distance matrix as the smoothing parameter. 

The polynominal function is defined as,
\begin{equation}
    \mathbf{k}(\mathbf{x},\mathbf{y}) = (\mathbf{x}'*\mathbf{y})^d
\end{equation}
where $d$ is the ploynominal degree. Here we adopted $d = 2$.
For both linear mapping and nonlinear mapping, the hyperparameters are generated in a certain range and the optimal hyperparameters are selected based on a metric defined using the estimated nICP error from the training datasets with cross validation. The kernalized nonlinear mapping function formulation is hereafter referred to as `Nonlinear Mapping with Gaussian Kernel' and `Nonlinear Mapping with Polynominal Kernel' for illustration in the paper.

\subsection{Noninvasive Features}
\label{SsecNoninvasiveFeatures}
In this section, the algorithm to compute noninvasive features is introduced in detail. Four types of feature vectors were computed and concatenated into one vector as input to the mapping-functions. The first type of feature vector is the impulse response features. The slow wave of ABP and R-R interval are selected as the inputs, the slow wave of CBv is treated as the output. Slow waves in ABP, and CBv can be described by oscillatory modes between 0.013 and 0.155 Hz \cite{MartinezCzosnyka2021}. The impulse response constructed from the model identified using slow-wave signals is treated as an impulse-response feature vector. The input signals are the slow waves from ABP and the R–R interval. The output signal is the slow wave from CBv. The second type of feature vector contains the latency features. The latency features are extracted from ABP and CBFV signals by computing the latency time between the onsets and the R peak in the ECG signal. The third type of feature vector is the cerebral arterial compliance features. The cerebral arterial compliance features are extracted by fitting the first derivative of ABP versus CBv on a pulse-by-pulse fashion. 

The fourth type of feature vector is the morphological clustering and analysis of continuous intracranial pressure (MOCAIP) feature vector \cite{HuXu2010}, \cite{HuXu2008}. The image on the right-hand side of Fig. \ref{FigZhao4} shows the landmarks on a CBv pulse and the table shows the MOCAIP feature vector was extracted as the 128 CBv pulse morphological metrics. It has been demonstrated that MOCAIP can be applied to the CBv signal. These metrics of the CBv pulses were reliably extracted using our MOCAIP algorithm \cite{KimBergsneider2011}, \cite{KimHu2011}. The MOCAIP algorithm automates the analysis of CBv pulses by designating the locations of three sub-peaks and valleys in the waveform as shown in Fig. \ref{FigZhao4}. We demonstrated that these metrics can be used as features to detect elevated ICP \cite{XieZhang2013}. In this paper, the concatenated feature vector is adopted as input to the mapping-function to evaluate the performance of LDM for nICP estimation. 

\subsection{Pre- and Post-Processing}
To further investigate the performance of the learned mapping function, pre- and post-processing were applied to the identification systems, input signals, and estimated mean nICP. Table \ref{Zhao.t1} summarizes the different scenarios in nICP estimation. `Normalization' refers to applying the Min-Max normalization method to the input signals prior to nICP evaluation. `Median value' and 'Mean value' stand for selecting the median and mean values of the mean nICP from the five identified systems with the smallest estimated nICP error by their mapping functions, respectively. `\ding{51}' indicates the inclusion of that computation. `\ding{55}' indicates the exclusion of that computation.

\begin{table}[h!] % Float environment
	\centering % Center the table
	\caption{Summary of various scenarios in nICP estimation.}
	\footnotesize
	\begin{tabular}{lccc} % Tabular environment
		\hline
		Scenario    & Normalization     & Median value & Mean value \\\hline\hline
		N0             & \ding{55}      &   \ding{55}    &  \ding{55}\\
		N0-Med         & \ding{55}      &   \ding{51}    &  \ding{55}\\
		N0-Mean        & \ding{55}      &   \ding{55}    &  \ding{51}\\\hline
		N1             & \ding{51}      &   \ding{55}    &  \ding{55}\\
		N1-Med         & \ding{51}      &   \ding{51}    &  \ding{55}\\
		N1-Mean        & \ding{51}      &   \ding{55}    &  \ding{51}\\\hline
		\hline
	\end{tabular}
	\label{Zhao.t1} % Optional: for referencing
\end{table}

\section{Data Preprocessing}
\label{sec:dataprocessing}
This section introduces the data collection and preprocessing for noninvasive intracranial pressure estimation. We adopted datasets from various locations to develop a comprehensive database of multimodality monitoring signals for brain-injured patients. The database includes electrocardiogram (ECG), ABP, CBv, and ICP signals, collected from the University of California, Los Angeles (UCLA), University of California, Davis (UCD), Emory University (IRB: STUDY00004039), University of São Paulo (USP) Medical School Hospital (IRB: NCT03144219), NeuraSignal, Inc (IRB: 14430), and Wroclaw Medical University (WMU) in Poland (IRB:
KB-620/2020, KB-133/2023). The data collection protocols were approved by the Institutional Review Boards of Emory University, and the data from these institutions were obtained under Emory-approved Data Use Agreements (DUAs).

\begin{table*}[ht!] % Float environment
	\centering % Center the table
	\caption{Summary of the datasets from various locations.}
	\scriptsize
	\begin{tabular}{lllllll} % Tabular environment
		\hline
		Datasets   & Number of Patients & ICP Monitoring Type & Range of ICP (mmHg) & Standard Deviation (mmHg) & Mean ICP (mmHg) & Median ICP (mmHg)\\\hline\hline
		UCLA  & 75  & N/A & [-29.06, 62.61]  & 8.24 & 14.28 & 13.39\\
		UCD   & 25  & EVD, parenchymal monitor&  [-7.57, 58.85]  & 5.65 & 13.35 & 13.32\\
		USP  & 35   & Intraventricular catheter&  [-1.52, 37.33]   & 6.15 & 13.74 & 12.59\\
		WMU     & 19   & Intraparenchymal probe&  [-7.72, 36.94]   & 7.15 & 10.32 & 9.02\\
		NeuraSignal & 1    & Intraventricular catheter &  [5.92, 26.07]    & 3.63  & 14.73 & 14.66\\
		Emory       & 1   & EVD &  [5. 52, 45.23]     & 2.69  & 10.25 & 9.29\\
		Total       & 156  & N/A &  [-29.06, 62.61]  &  6.75 & 13.36 & 12.97\\
		\hline\hline
        \multicolumn{5}{l}{\footnotesize \,EVD: External ventricular drain.} 
	\end{tabular}
	\label{Zhao.t2} % Optional: for referencing
\end{table*}

\begin{table*}[ht!] % Float environment
	\centering % Center the table
	\caption{Patient demographics and pathology distribution.}
	\scriptsize
	\begin{tabular}{llll} % Tabular environment
		\hline
		Category   & Level & Number & Percentage (\%)\\\hline\hline
		Age (years)  & Mean & 44.50  & —\\
                     & Median & 42 & —\\\hline
		Age group   & 18-39  &  171  &  40.33 \\
                    & 40-59  &  118  & 27.83 \\
                    & 60-79  &  123  & 29.01 \\
                    & $\geq$80 & 5 & 1.18\\\hline
		Sex  & Female   &  164   &  38.68   \\
             & Male     &  259    &  61.08  \\\hline
		Pathology / Diagnosis & Traumatic brain injury (TBI)  &  118    &  27.83   \\
        & Subarachnoid hemorrhage (SAH) & 101 & 23.82\\
        & Other & 205 & 48.35\\
		\hline\hline
	\end{tabular}
	\label{Zhao.t3} % Optional: for referencing
\end{table*}

There are 156 patients in total, with 75 from UCLA \cite{KimHamilton2012}, 25 from UCD, 35 from USP Medical School Hospital, 19 from WMU, 1 from NeuraSignal, and 1 from Emory University. The data selection process is a critical step in preparing datasets for analysis or model development. To ensure compatibility with the algorithm developed, in this study, the data selection process involves evaluating the completeness of the data, ensuring the quality of the data, and selecting valid features. Consequently, the number of patients across locations differs according to subject enrollment and data quality. A summary of the datasets information can be found in Table \ref{Zhao.t2}. The main ICP is distributed in the range $[-29.06, 62.61]\hspace{1mm}$mmHg from brain-injured patients. The majority of the ICP values are concentrated within the range of $[12, 14]$\hspace{1mm}mmHg with the mean ICP as $13.36\hspace{1mm}$mmHg. The datasets were divided into segments of 360 heartbeats per entry, resulting in a total of 448 entries. Note that the ICP was collected using various monitoring types and at different measurement locations; this may introduce hydrostatic offsets that affect the estimation results. The demographics and pathology distribution of the entries are shown in Table \ref{Zhao.t3}. Note that demographic information is not available for all entries; therefore, we summarize demographics based on 424 entries. 

The 360 heartbeats length is chosen as a tradeoff between providing a sufficient number of signal samples for system identification and using a short enough signal segment to avoid violation of the linearity assumption that is used for using a linear dynamic model to model coupled CBF and CSF circulatory systems. To test the performance of the machine learning algorithm, we separated the datasets into training and testing datasets based on patient ID. There are 357 entries randomly selected as training datasets, the rest 91 entries are adopted as testing datasets. For the training datasets, the ICP is distributed in the range of $[-17.81, 60.06]\hspace{1mm}$mmHg. For the testing datasets, the ICP is distributed in the range of $[-29.06, 69.80]\hspace{1mm}$mmHg. Among the 448 entries, there are 310 entries with ECG signal and 138 entries without ECG signal. For the datasets with ECG, the ECG signal is adopted to identify peaks for segmentation of other physiological signals. For datasets without ECG, segmentation is performed using the IMS algorithm \cite{KarlenAnsermino2012} on ABP signal, as the ABP signal offers the highest quality. Then the segmentation of ICP and TCD is performed based on their latency time relative to the ABP signal. The onsets of ICP and TCD are searched in the nearest time window based on the onsets of ABP signal. The relative time search window for TCD and ICP signals is set to $0.5s$ before the onset of the ABP signal and $0.15s$ after it. Here the relative time is determined through trial and errors to guarantee the quality of signal segmentation. 

\begin{figure*}[ht]
	\centering
	\includegraphics[width=1.0\textwidth]{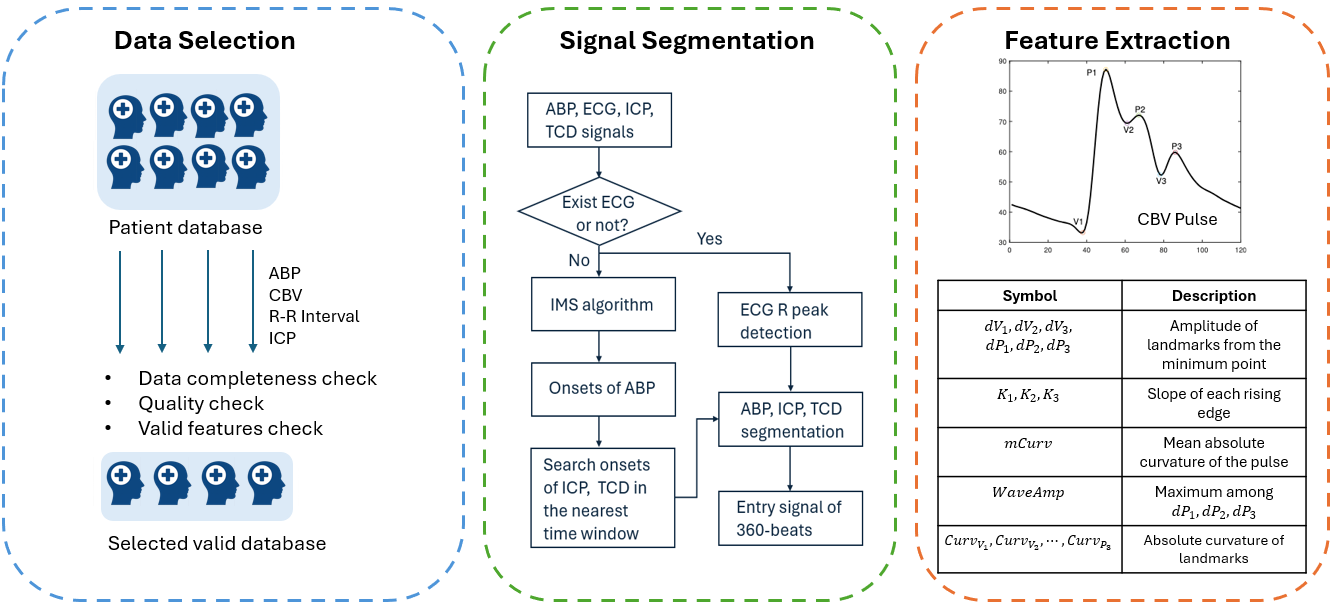} % Adjust width as needed
	\caption{Data processing and feature extraction procedures for nICP estimation. Left: Data selection procedure from the database. Middle: Signal segmentation for signals with and without ECG. Right: Feature extraction for CBv pulse.}
	\label{FigZhao4}
\end{figure*}

Fig. \ref{FigZhao4} shows the data processing in general, including the data selection, signal segmentation, and feature extraction. The 448 entries were selected using the following procedures: (1) data completeness and quality check and (2) feature validity check. After the data completeness and quality check, signals from each recording were segmented into entries of 360 beats using the MOCAIP toolbox. Each recording is at least 360 beats long. Each entry was then manually reviewed by verifying the validity of the extracted MOCAIP features. There are 128 features in total, and some representative features are shown on the right-hand side of Figure 4. A flowchart is created as shown in the middle of Fig. \ref{FigZhao4} to show the segmentation for datasets with ECG and without ECG to obtain the individual entry. All datasets, either non-invasive or invasive, are collected synchronously and resampled at 400 Hz. For entries with ECG, the R-R interval is computed using a general biomedical signal processing toolbox for QRS detection. For entries without ECG, the R-R interval is determined by detecting the onsets of the ABP signal, which serves as a surrogate QRS R peak position. Table \ref{Zhao.t4} gives a summary of the training and testing datasets in terms of the number of entries, number of patients, ECG signal availability. Note that there are 4 testing entries and 3 training entries with a mean ICP below 0 $\hspace{1mm}$mmHg, which could be due to the measuring position and artifacts. 

\begin{table*}[ht!] % Float environment
	\centering % Center the table
	\caption{Summary of various scenarios in nICP estimation.}
	\scriptsize
	\begin{tabular}{lcccc} % Tabular environment
		\hline
		Datasets       & Number of entries  & Number of patients & Entries w/ ECG\textsuperscript{a} & Entries w/o ECG\textsuperscript{b} \\\hline\hline
	    Training        & 357        &  114      &   274  &  83  \\
		Testing         & 91           &  42       &    36   &  55\\  
		\hline
        \multicolumn{5}{l}{\footnotesize \textsuperscript{a}\,Entries w/ ECG: Number of entries with ECG signals. 
        \textsuperscript{b}\,Entries w/o ECG: Number of entries without ECG signals.} 
	\end{tabular}
	\label{Zhao.t4} % Optional: for referencing
\end{table*}

\section{Results}
\label{sec:results}

This section shows the results of noninvasive intracranial pressure estimation. Section \ref{SsecResultsPreTrained} presents the results of the trained model, evaluated using both the training and testing datasets.

\subsection{Results from Trained Mapping Functions}
\label{SsecResultsPreTrained}
The Bland-Altman plot is adopted to show the performance of nICP estimation. Bland-Altman analysis is a common tool adopted in clinical studies to show the agreement between two quantitative methods of measurement \cite{Giavarina2015}. The Bland-Altman plot shows the mean value of the y axis and limits of agreement. Limits of agreement (LOA) is defined as mean $\pm$ 1.96 $\times$ the standard deviation of the mean value. Multiple scenarios in Table \ref{Zhao.t1} are selected to illustrate the estimation accuracy and shown in Table \ref{Zhao.t5}. Table \ref{Zhao.t5} summarizes the results from computation scenario N1-Mean for testing entries with linear and nonlinear mappings. It has been observed that the results from N0 and N1 are identical. Similarly, the results from N0-Median and N1-Median are identical, as are the results from N0-Mean and N1-Mean. The averaged value from N1, N1-Median, and N1-Mean is aslo adopted to show the estimation results in Table \ref{Zhao.t5}. The results indicate that normalization exert only a minor influence on the nICP estimation errors. The percentage of estimation errors within 2$\hspace{1mm}$mmHg and between 2$\hspace{1mm}$mmHg and 6$\hspace{1mm}$mmHg are also calculated to demonstrate the accuracy of the estimation results. 

Table \ref{Zhao.t5} also shows the nICP estimation results from testing entries with linear mapping function without ranking constraints. In average, there are 20.16\% of entries with mean estimation error below 2$\hspace{1mm}$mmHg, 23.08\% of the estimation errors between 2$\hspace{1mm}$mmHg to 6$\hspace{1mm}$mmHg. In total, there is 43.24\% below 6$\hspace{1mm}$mmHg. For the linear mapping function with ranking constraints, there is an average increase of 1.83\% in the number of entries with nICP estimation errors below 2$\hspace{1mm}$mmHg, and a 22.35\% increase in entries with errors between 2$\hspace{1mm}$mmHg and 6$\hspace{1mm}$mmHg. Overall, N1-Median demonstrates the highest accuracy with 69.25\% below 6$\hspace{1mm}$mmHg. Compared to the linear mapping function without constraints, incorporating ranking constraints significantly reduces the nICP estimation error, resulting in a 24.18\% increase in the number of entries with errors below 6$\hspace{1mm}$mmHg.

\begin{table*}[ht!] % Float environment
	% \begin{table}[!ht] % Float environment
	\centering % Center the table
	\caption{Summary of nICP estimation mean errors below $2\hspace{1mm}$mmHg, and between $2\hspace{1mm}$mmHg and $6\hspace{1mm}$mmHg from testing entries for linear mappings and nonlinear mappings with N1-Median and averaged value from various scenarios.}
    \scriptsize
	\begin{tabular}{c|cclll} % Tabular environment
		\hline
		  Algorithms & Evaluation level & Scenarios      & $ E<2\hspace{1mm}$mmHg 
          $(\%)$ & $ 2\hspace{1mm}$mmHg $< E < 6\hspace{1mm}$mmHg $(\%)$ & $ E<6\hspace{1mm}$mmHg 
          $(\%)$ \\\hline\hline
		Linear Mapping without Constraints & Entry level  &  N1-Median    & 25.28  &  19.79 & 45.07\\
        & & Average        & 20.16  &  23.08 & 43.24\\\cline{2-6}
        & Patient level & N1-Median & 33.33 & 26.19 & 50.00\\
        &               & Average & 28.58 & 30.95 & 50.00\\\hline
		Linear Mapping with Constraints & Entry level & N1-Median    & 21.99  &  47.26 & 69.25\\
        & &  Average        & 21.99  &  45.43 & 67.42\\\cline{2-6}
        & Patient level & N1-Median & 30.95 & 52.38 & 71.43\\
        &               & Average & 23.81 & 54.76 & 69.05\\\hline
		Nonlinear Mapping with Gaussian Kernel & Entry level & N1-Median    & 31.88  &  34.07 & 65.95\\
        & & Average        & 32.24  &  35.53 & 67.81\\\cline{2-6}
        & Patient level & N1-Median & 35.71 & 45.24 & 69.05\\
        &               & Average & 35.71 & 47.62 & 71.43\\\hline
		Nonlinear Mapping with Polynominal Kernel & Entry level & N1-Median    & 30.78  &  35.17 & 65.95\\
        & & Average        & 31.51  &  36.27 & 67.78\\\cline{2-6}
        & Patient level &  N1-Median & 38.10 & 45.24 & 69.05\\
        &               & Average & 38.10 & 45.24 & 71.43\\\hline 
	\end{tabular}
    \label{Zhao.t5}
\end{table*}

\begin{table*}[ht!] % Float environment
	% \begin{table}[!ht] % Float environment
	\centering % Center the table
	\caption{Summary of nICP estimation errors using mean, standard deviation and median metrics from testing entries for linear mappings and nonlinear mappings.}
    \scriptsize
	\begin{tabular}{c|cclll} % Tabular environment
		\hline
		  Algorithms & Evaluation level & Mean  &  STD & Median\\\hline\hline
		Linear Mapping without Constraints & Entry level  &  7.99    & 45.1  &  5.63 \\\hline
		Linear Mapping with Constraints & Entry level & -0.93    & 7.14  &  -1.52 \\\hline
		Nonlinear Mapping with Gaussian Kernel & Entry level & 0.073    & 6.98  &  0.15 \\\hline
        \multicolumn{5}{l}{\footnotesize \,STD: Standard deviation.}
	\end{tabular}
    \label{Zhao.t6}
\end{table*}

Based on the results from testing entries with nonlinear mapping function using various kernels, including the Gaussian kernel and the polynomial kernel. The nonlinear mapping with a Gaussian kernel shows the highest accuracy improvement.
Compared to linear mapping without constraints, there is 24.57\% increment entries in terms of nICP estimation error below 6$\hspace{1mm}$mmHg.

Table \ref{Zhao.t5} also summarizes the mean nICP estimation errors for testing entries using linear mapping with and without constraints, as well as nonlinear mapping, at the patient level. At the patient level, the estimation error is calculated based on the patient ID, using the mean estimation error across various entries from the same patient. At the patient level, linear mapping without constraints achieved clinically acceptable accuracy (error below 2$\hspace{1mm}$mmHg) \cite{EvensenEide2020} in average only 28.58\% of the patients. 30.95\% of the patients showed moderate accuracy (between 2$\hspace{1mm}$mmHg to 6$\hspace{1mm}$mmHg). Note that some patients appear in both ranges, resulting in approximately 43.24\% of the patients having estimation errors below 6$\hspace{1mm}$mmHg. 

By incorporating the constraints, there is 4.77\% decreasing of the patients have estimation errors less than 2$\hspace{1mm}$mmHg, 23.81\% improvements in the range of 2$\hspace{1mm}$mmHg to 6$\hspace{1mm}$mmHg, and 21.43\% improvements below 6$\hspace{1mm}$mmHg. Same for the results computed from the nonlinear mapping with Gaussian and polynomial kernels, there are more than 20\% improvements for estimation results below 6$\hspace{1mm}$mmHg.

Table \ref{Zhao.t6} shows the summary of nICP estimation errors using mean, standard deviation, and median metrics from testing entries for
linear mappings and nonlinear mappings. It is observed that the nonlinear mapping with a Gaussian kernel yields the minimum values in terms of mean estimation error, standard deviation, and median error.

\begin{figure*}[ht]
	\centering
	\begin{subfigure}{0.8\textwidth}
        \includegraphics[width=\linewidth]{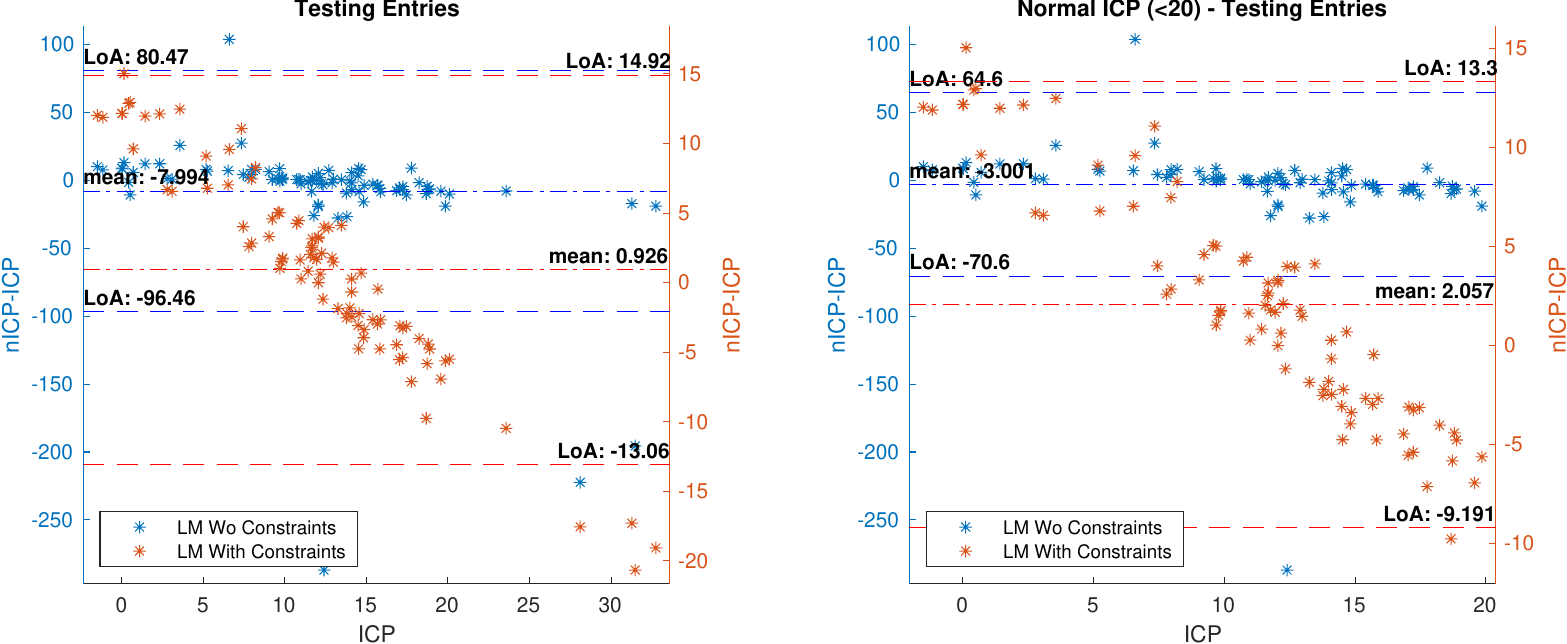}
        \caption{}
    \end{subfigure}
    \vfill
    \begin{subfigure}{0.8\textwidth}
        \includegraphics[width=\linewidth]{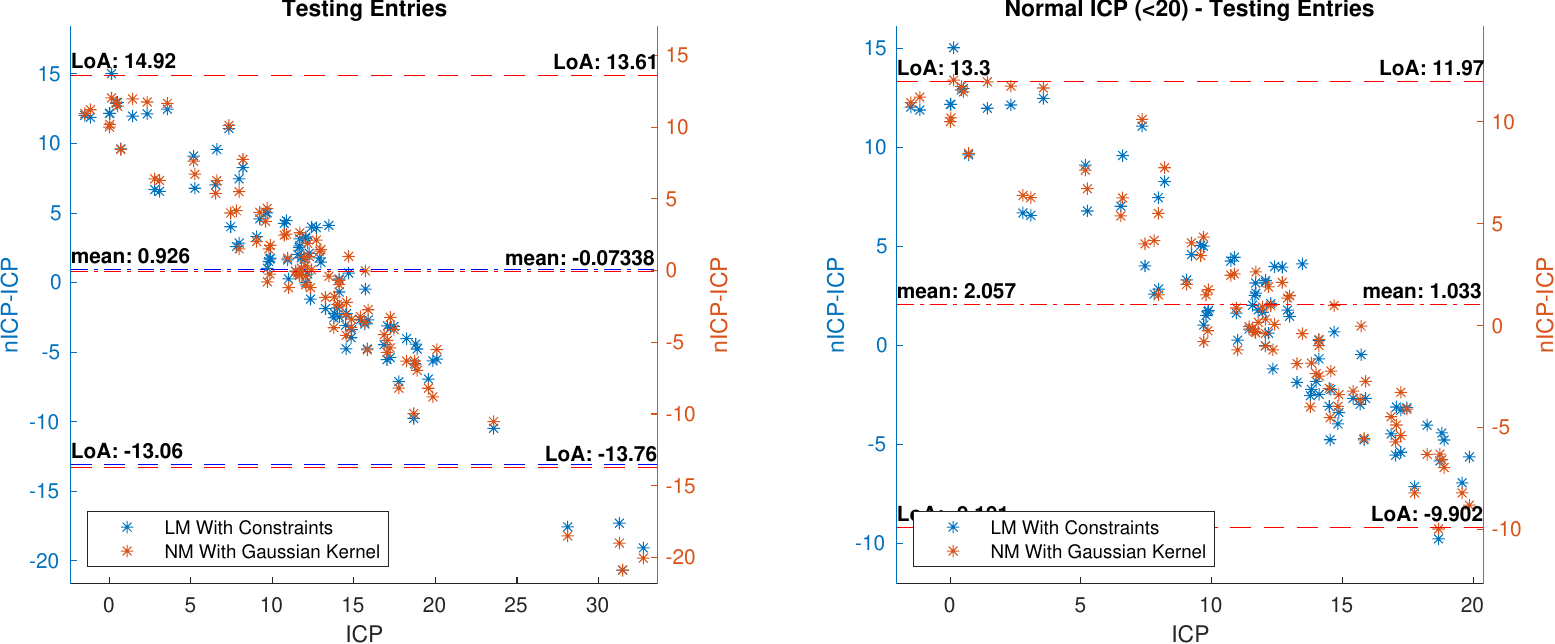}
        \caption{}
    \end{subfigure}
	\caption{Bland–Altman plots for testing entries with comparisons from different algorithms. (a): Bland-Altman plot for nICP results comparions between the linear mapping without constraints and linear mapping with constraints. (b): Bland-Altman plot for nICP results comparions between the linear mapping with constraints and nonlinear mapping. Values on the left y-axis represent the results from data points in blue. Values on the right y-axis in the images represent the results from data point in orange. `LM Wo Constraints' stands for the linear mapping without constraints. `LM With Constraints' stands for the linear mapping with constraints. `NM With Gaussian Kernel' stands for nonlinear mapping with Gaussian kernel.}
	\label{FigZhao5}
\end{figure*}

Fig. \ref{FigZhao5} presents the nICP estimation results using Bland–Altman plots from testing entries for the original linear mapping without constraints, linear mapping with constraints, and nonlinear mapping, including all testing entries as well as those with normal ICP. Normal ICP is defined as a ground truth ICP value below 20$\hspace{1mm}$mmHg. Specifically, the N1-Median algorithm is selected to demonstrate the accuracy of the results. Each dot in the figure stands for the mean estimation error from each entry. For linear mapping without constraints, the mean estimation error is around -8$\hspace{1mm}$mmHg for testing entries, for normal ICP below 20$\hspace{1mm}$mmHg the mean estimation error is about -3$\hspace{1mm}$mmHg. Moreover, numerous estimation outliers are observed in the unconstrained linear mapping. We observed that the mean estimation error is around 1$\hspace{1mm}$mmHg for testing entries from linear mapping with constraints, indicating good estimation performance once the ranking constraints are included in the linear mapping. For entries with normal ICP the mean estimation error is around 2$\hspace{1mm}$mmHg. The trained mapping function performs well for entries with normal ICP.

\begin{figure}[ht]
	\centering
	\includegraphics[width=0.65\textwidth]{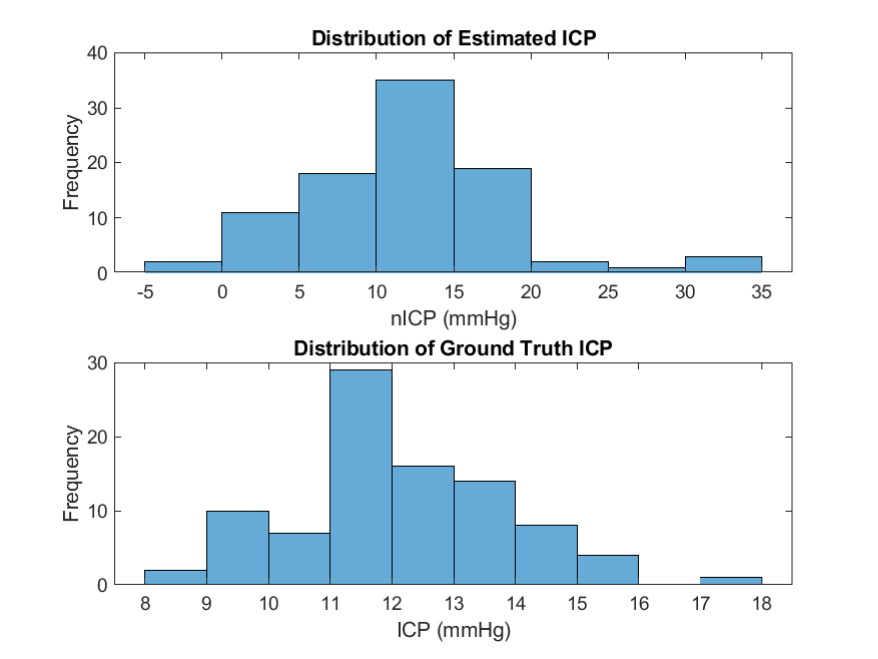} % Adjust width as needed
	\caption{Distribution of estimation ICP and ground truth ICP from nonlinear mapping with Gaussian kernel.}
    \label{FigZhao5.3}
\end{figure}

Fig. \ref{FigZhao5} also shows the testing entries with N1-Median from nonlinear mapping. From the results, we observed that for all testing entries, it has the minimum mean estimation error as around -0.1$\hspace{1mm}$mmHg. The results demonstrate that, overall, the nonlinear mapping yields a smaller mean estimation error across all subjects, accompanied by higher confidence. Fig. \ref{FigZhao5.3} shows the distribution comparison between the estimated ICP and the ground truth ICP from nonlinear mapping with Gaussian kernel. 

\begin{figure}[ht]
	\centering
	\includegraphics[width=0.6\textwidth]{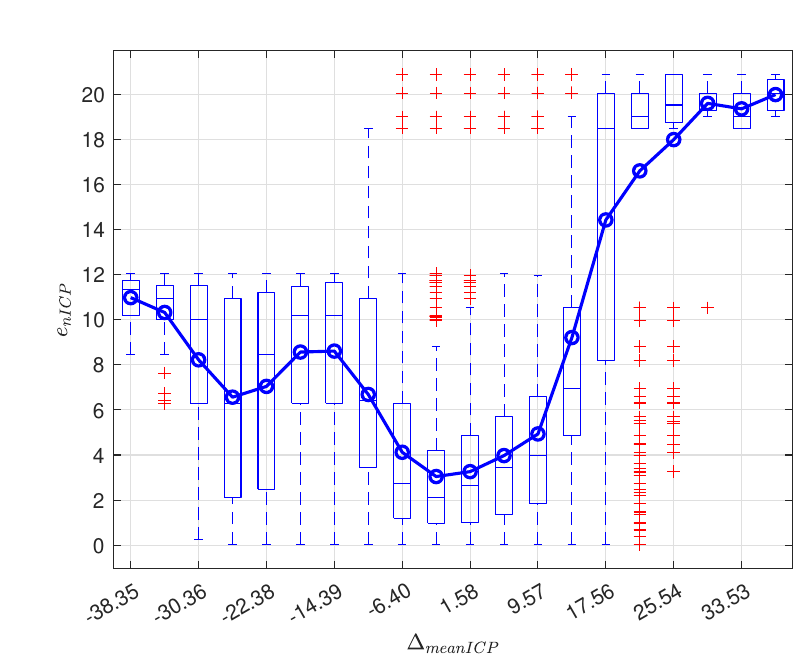} % Adjust width as needed
	\caption{Box plot of the nICP estimation error in terms of the mean ICP difference between the testing entries and the training entries. `o' stands for the mean value of each bin.}
    \label{FigZhao6}
\end{figure}

\begin{figure*}[ht]
       \centering
    \begin{subfigure}{0.4\textwidth}
        \includegraphics[width=\linewidth]{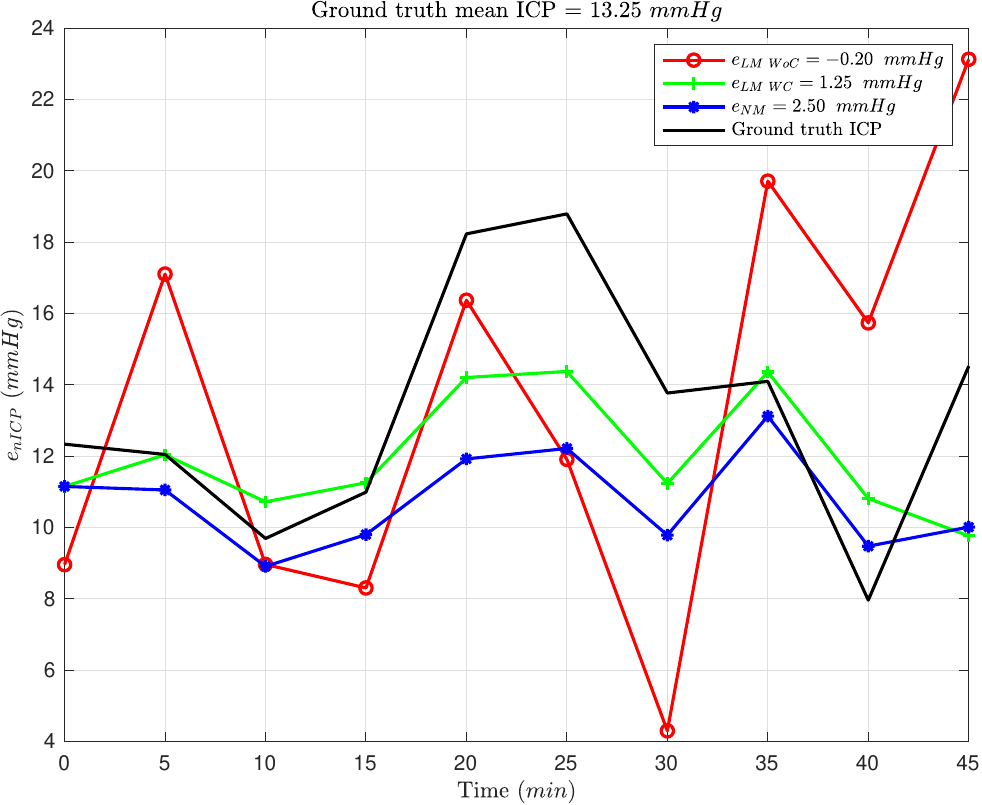}
    \end{subfigure}
    \hspace{0.5cm}
    \begin{subfigure}{0.4\textwidth}
        \includegraphics[width=\linewidth]{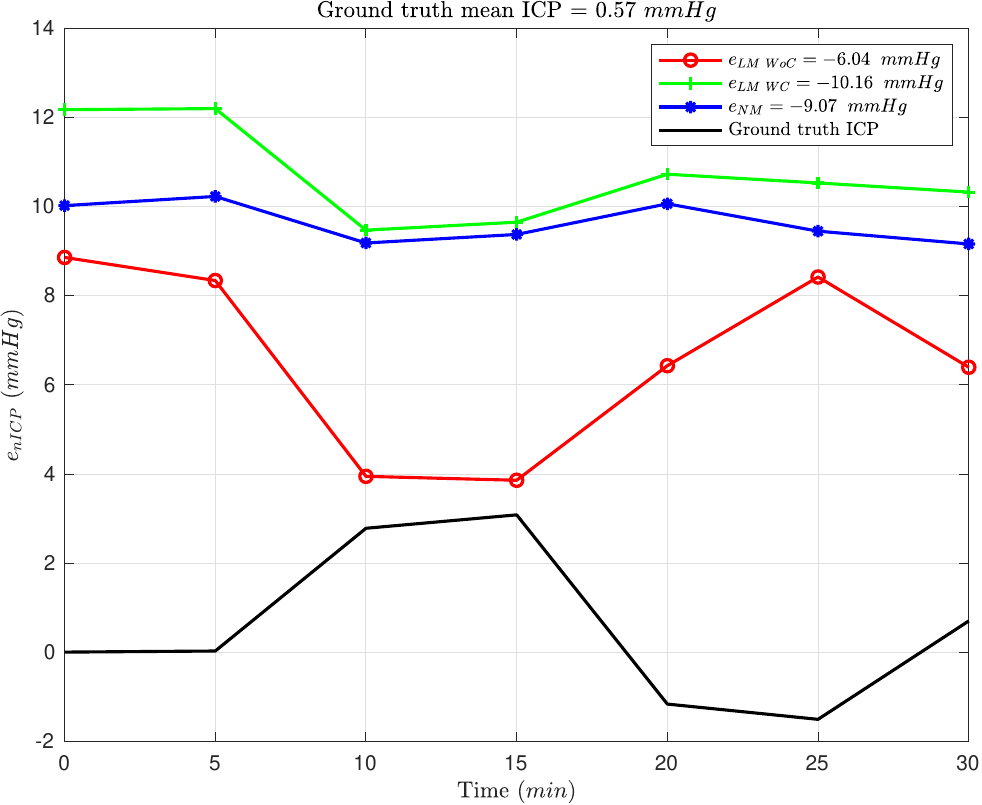}
    \end{subfigure}
    \vfill
    \begin{subfigure}{0.4\textwidth}
        \includegraphics[width=\linewidth]{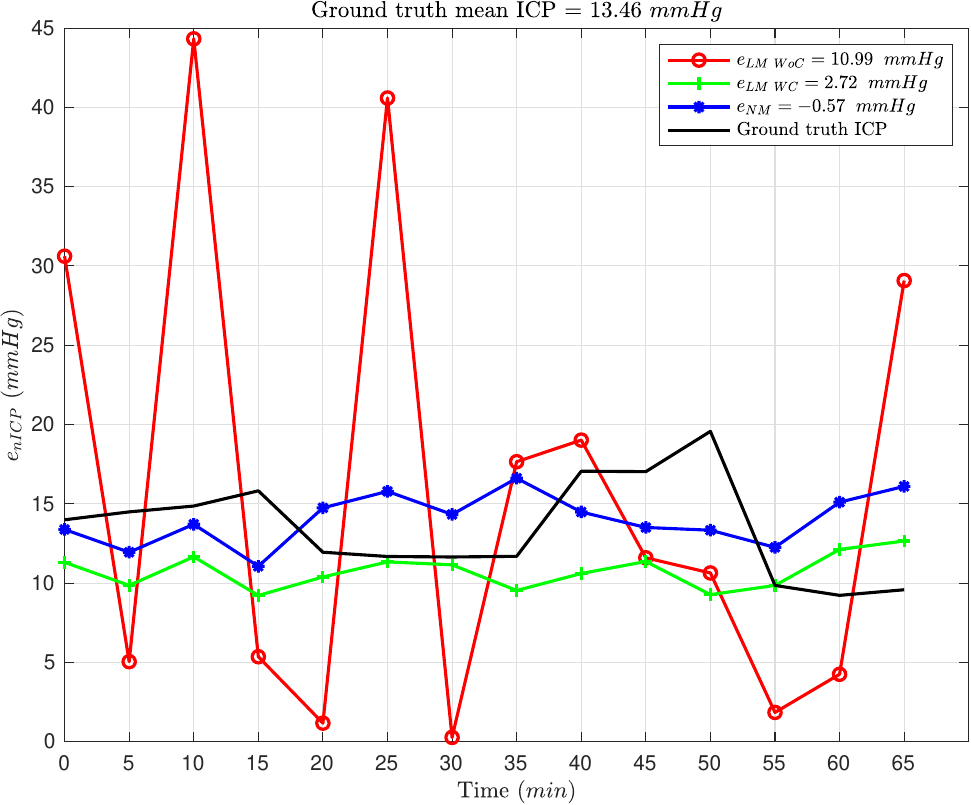}
    \end{subfigure}
    \hspace{0.5cm}
    \begin{subfigure}{0.4\textwidth}
        \includegraphics[width=\linewidth]{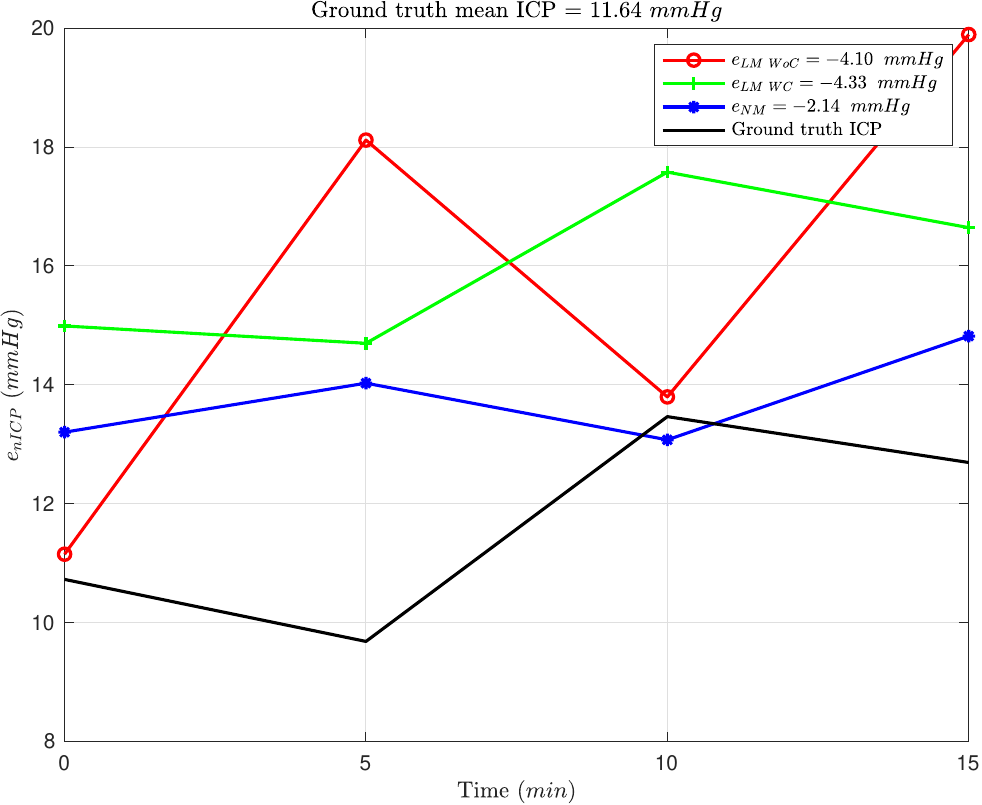}
    \end{subfigure}
    \caption{Error plots for various testing entries with N1-Median from different algorithms. From left to right: 
    (1): testing entry with ground truth mean ICP = $13.25\hspace{1mm}$mmHg, 
    (2): testing entry with ground truth mean ICP = $0.57\hspace{1mm}$mmHg. 
    (3): testing entry with ground truth mean ICP = $13.46\hspace{1mm}$mmHg, 
    (4): testing entry with ground truth mean ICP = $11.64\hspace{1mm}$mmHg. $e_{LM\hspace{1mm}WoC}$ stands for the estimation error from linear mapping without constraints. $e_{LM\hspace{1mm}WC}$ stands for the estimation error from linear mapping with constraints. $e_{NM}$ stands for the estimation error from nonlinear mapping.}
    \label{FigZhao7}
\end{figure*}

Fig. \ref{FigZhao6} shows the nICP estimation error as a function of the mean ICP difference between the testing entries and the training entries. The objective is to investigate the influence of the ICP value differences on the training and testing datasets. It is observed that the mean ICP difference between the training and testing entries has a significant influence on the nICP error, with the minimum estimation error occurring around $\Delta_{meanICP} = 1.58\hspace{1mm}$mmHg. Fig. \ref{FigZhao7} presents the estimated ICP values from various patients over a prolonged time period using different algorithms. Patients with normal and low ICP are used as examples to illustrate the estimation results. The results indicate that linear mapping without constraints yields the largest estimation error for patients with normal ICP over a long time period as indicated by the patient case with ground truth ICP as $13.46\hspace{1mm}$mmHg, while nonlinear mapping produces the smallest estimation error.

\subsection{Ranking Constraints Analysis}
\label{SsecResultsRankingConstraints}
\begin{figure*}[ht]
    \centering
    % ---- First row ----
    \begin{subfigure}{0.45\textwidth}
        \includegraphics[width=\linewidth]{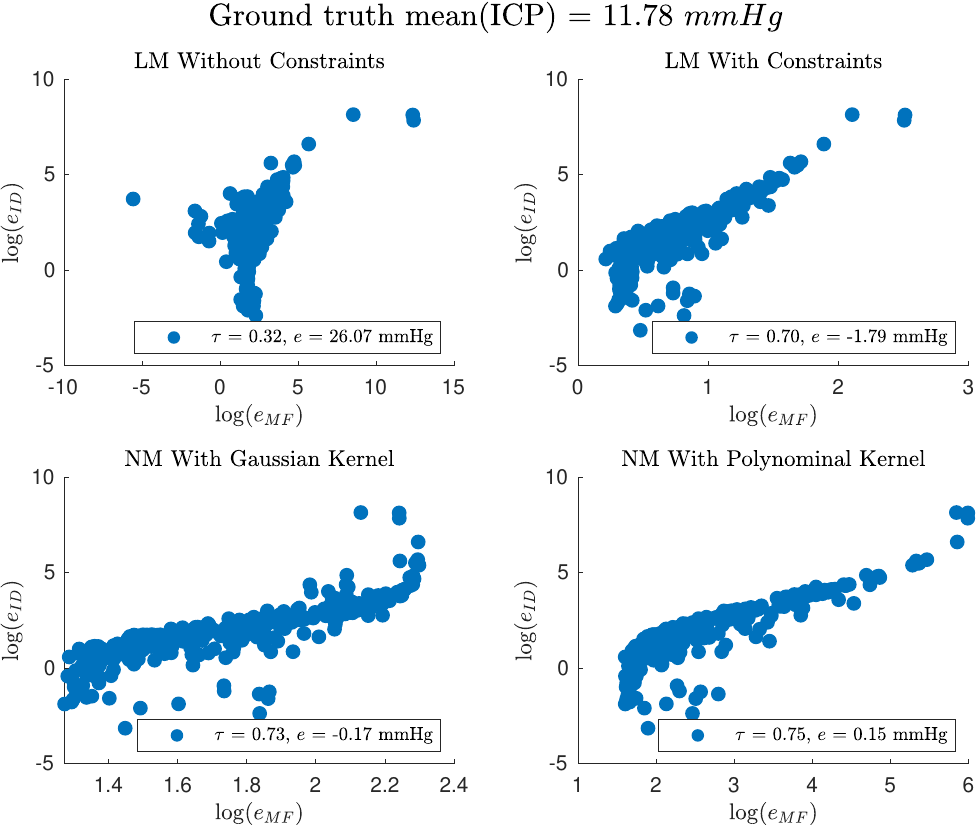}
    \end{subfigure}
    \begin{subfigure}{0.45\textwidth}
        \includegraphics[width=\linewidth]{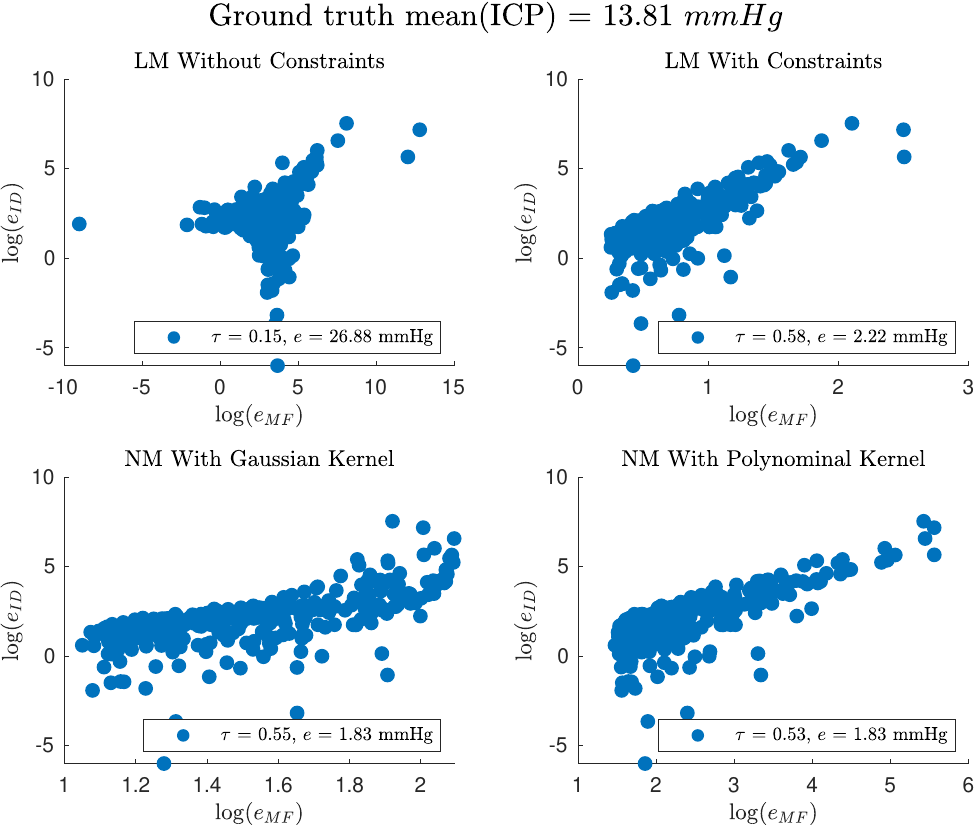}
    \end{subfigure}

    \caption{Error plots for various testing entries with N1-Median, illustrating the performance achieved by incorporating the constraints. `LM' stands for `Linear Mapping', `NM' stands for `Nonlinear Mapping'. $e_{ID}$ stands for the errors from the identified models. $e_{MF}$ stands for the errors from the mapping functions.}
    \label{FigZhao8}
\end{figure*}

Fig. \ref{FigZhao8} illustrates the Kendall correlation between the errors from the mapping functions and those from the identified systems. For ground truth mean ICP values within the normal range (e.g., $11.78\hspace{1mm}$mmHg or $13.81\hspace{1mm}$mmHg), the nonlinear mappings using polynomial and Gaussian kernels achieve comparable Kendall correlation coefficients with smaller errors, whereas the linear mapping without constraints produces the lowest correlation and the highest estimation error. This finding indicates that incorporating ranking constraints is highly effective for estimating normal ICP, yielding improvements of at least $90\%$.

\section{Discussion}
\label{sec:discussion}

 Our solution for a practical and potential nICP technology creatively integrates machine learning and dynamic system identification to provide individualized calibration. Individualization is realized by a novel idea based on learning ``mapping functions'' that estimate nICP prediction errors associated with identified linear dynamic models (LDMs) within a pre-established database. Ranking available LDMs based on the estimated nICP errors determines the optimal model to simulate ICP for individual cases. When the case database lacks cases representative of a test instance, the optimal LDM may produce inaccurate nICP estimation. Learning the mapping-functions was challenging until a recent breakthrough. This breakthrough incorporates novel algorithmic components that go beyond conventional neural networks based machine learning methods. In particular, this mapping-function learning problem cannot be solved by simply using off-the-shelf machine-learning techniques. Instead, problem-specific information needs to be incorporated to find a solution. Our past efforts used algorithms such as linear regression \cite{PrunetAsencio2012}, support vector machines \cite{RazumovskyArmonda2011}, and spectral regressions \cite{HuXu2010} to independently estimate mapping-function for each LDM. Such efforts were met with moderate success because ranking constraints were not considered in the learning process. Incorporating ranking constraints, however, presents a practical challenge. Therefore, we designed an approximate but computationally efficient solution. To our knowledge, this paper will be the first to adopt a bespoke machine learning algorithm specifically designed to address the key challenges inherent in noninvasive ICP estimation.

We also propose to use morphological metrics derived from CBv pulse as input features to mapping-functions. While the influence of ICP on CBv pulse morphology is well known \cite{DeDe2002}, most existing approaches for relating CBv to ICP are focused on the single pulsatility index (PI). PI only provides amplitude information of CBv pulses, and its performance for nICP assessment is inconsistent  \cite{CuturiDoucet2011}, \cite{GonenAlpaydın2011}. In contrast, this study adopts the MOCAIP algorithm \cite{AggarwalBrooks2008} for CBv signal analysis. In addition to CBv pulse morphological metrics, impulse response features extracted from the R-R interval, ABP, and CBv, as well as the latency and cerebral compliance features, are adopted as input features to the mapping functions.

A database of multimodality monitoring signals from 156 patients across six institutions is built as the training and testing database. The trained mapping function shows promising nICP estimation accuracy for normal ICP ranges, achieving errors within 6$\hspace{1mm}$mmHg for approximately 69\% of measurements. The results also indicate that our approach is grounded in the physiological dynamic relationships among CBv, ABP, and R-R interval, which are common in brain conditions, leading to reasonable intracranial pressure estimation under specific conditions. 

However, our analysis reveals a critical limitation: the framework systematically underestimates elevated ICP ($>$20$\hspace{1mm}$mmHg) with a mean error of -15.74$\hspace{1mm}$mmHg for nonlinear mapping and -15.09$\hspace{1mm}$mmHg for linear mapping with constraints. For entries with hypertension, the model tends to underestimate the ICP, as most of the ICP values in the training data are distributed within the range of $12-14\hspace{1mm}$mmHg. It turns out that a few entries are distributed within the high ICP range, which leads to inaccurate nICP estimation. Additionally, data quality assessment remains an important challenge in the feature extraction of the signals, especially for the TCD signal. Whether TCD signals are effective for noninvasive monitoring of intracranial pressure is still under debate \cite{CardimRobba2016}. The TCD signal is highly sensitive to variations caused by probe placement and must be handled by well-trained technicians, which can lead to noise and artifacts. Recent studies show that transcranial signal can also be obtained by using a conformal ultrasound patch \cite{ZhouGao2024}, which could support the development of noninvasive intracranial pressure estimation through improved hardware and potentially reduce operator dependence. The developed approach is not limited to TCD signals. It can also be applied to other noninvasive physiological signals that reflect ICP through brain hemodynamics, such as near-infrared spectroscopy (NIRS). In this study, we illustrate the feasibility of the method using TCD signals as a representative example, demonstrating how our proposed approach can extract ICP-related information from hemodynamic measurements. Besides the TCD signals, the type of ICP monitoring can also affect the final results, since mis-leveling can introduce hydrostatic offsets. Furthermore, optimization algorithms may need to be developed to select the optimal hyperparameters for improving the performance of nICP estimation. 

For clinical translation, a kernel-warping procedure could be developed to incorporate the input feature of a $de$ $novo$ instance to obtain a data-dependent kernel. Moreover, further investigation is needed to understand how patient-specific features influence the nICP estimation error. A calibration function can be developed to establish the relationship between the error from the mapping function and the true nICP estimation error to properly handle $de$ $novo$ instances in clinical practice. Further physiologically based constraints can also be considered during the system identification and mapping function training procedures. As demonstrated by physics-informed neural networks in physiological signal modeling and processing \cite{ZhaoFattahi2025}, incorporating physics constraints into the training process of machine learning models can significantly improve the performance of prediction.

\section{Conclusion}
\label{sec:conclusion}

This study introduces a novel framework for nICP estimation based on machine learning algorithms incorporating ranking constraints. Accurate and continuous estimation of intracranial pressure is crucial for managing patients with neurological disorders, however, invasive monitoring techniques remain the clinical standard, limiting their routine application due to associated risks. The proposed framework aims to bridge this gap by leveraging multimodal physiological signals and advanced learning strategies to achieve noninvasive estimation. For validation, we used an available multimodality monitoring database including ABP, CBFV, and invasive ICP signals from 156 patients across six institutions; while not exhaustive, it offers a useful starting point for assessment. The proposed method demonstrated promising performance, achieving certain estimation accuracy across diverse clinical conditions and institutions. Preliminary results confirm the feasibility and generalizability of the ranking-constrained learning approach for noninvasive ICP monitoring. Nevertheless, further research is required to refine model interpretability and clinical applicability, especially for patients with intracranial hypertension. Future work will focus on in-depth analysis of elevated ICP, extraction of additional morphological and physiological features, systematic hyperparameter optimization, and the development of patient-specific kernels to capture individual cerebrovascular dynamics. 

\bibliographystyle{unsrtnat}
\bibliography{nICPLibrary_SN.bib}  %%% Uncomment this line and comment out the ``thebibliography'' section below to use the external .bib file (using bibtex) .

\end{document}